\documentclass[10pt,twocolumn,letterpaper]{article}

\usepackage{wacv}              

\makeatletter
\@namedef{ver@everyshi.sty}{}
\makeatother
\usepackage{tikz}

\usepackage{amsthm}
\usepackage[utf8]{inputenc}
\usepackage{enumitem,amssymb,amsmath,pifont}
\usepackage{graphicx}
\usepackage{booktabs}
\usepackage{tabularx}
\usepackage{multirow}
\usepackage{adjustbox}
\usepackage{afterpage}

\usepackage{array}
\newcolumntype{H}{>{\setbox0=\hbox\bgroup}c<{\egroup}@{}}
\usepackage{float}
\usepackage[skip=5pt, belowskip=-10pt]{caption}

\newtheoremstyle{mydefinitionstyle}
  {12pt}
  {2pt}
  {}
  {}
    {\bfseries}
  {.}
  {.5em}
  {\thmname{#1}\thmnumber{ #2}\thmnote{ [\textit{#3}]}}

\theoremstyle{mydefinitionstyle}
\newtheorem{mydefinition}{Definition}
\newtheorem*{mydefinition*}{}

\makeatletter
\newcommand*{\centerfloat}{%
	\parindent \z@
	\leftskip \z@ \@plus 1fil \@minus \textwidth
	\rightskip\leftskip
	\parfillskip \z@skip}
\makeatother

\def\wacvPaperID{1123} 
\def\confName{WACV}
\def\confYear{2024}

\usepackage[boldmath]{numprint}
\newlist{todolist}{itemize}{2}
\setlist[todolist]{label=$\square$}

\newcommand{\DC}{\texttt{DC}}%
\newcommand{\DU}{\texttt{VDU}}%
\newcommand{\ECE}{$\mathrm{ECE}$} 
\newcommand{\AURC}{$\mathrm{AURC}$} 
\DeclareMathOperator*{\argmax}{arg\,max}

\newcommand{\cmark}{\ding{51}}%
\newcommand{\xmark}{\ding{55}}%

\newcommand{\jordyskeleton}[1]{#1}

\newcommand{\commentout}[1]{\ignorespaces}

\usepackage{xargs}  
\usepackage[colorinlistoftodos,prependcaption]{todonotes} 
\definecolor{OliveGreen}{cmyk}{0.64,0,0.95,0.40}
\definecolor{ForestGreen}{RGB}{34,139,34}
\colorlet{darkgreen}{green!60!black!80!}
\colorlet{darkorange}{orange!60!black!80!}
\colorlet{purple}{red!80!green!20!blue!60!}
\newcommandx{\draft}[2][1=]{\todo[linecolor=ForestGreen,backgroundcolor=ForestGreen!25,bordercolor=ForestGreen,inline,caption={},#1]{#2}}
\newcommandx{\evidence}[2][1=]{\todo[linecolor=darkorange,backgroundcolor=darkorange!25,bordercolor=darkorange!75,inline,#1,noprepend,caption={}]{\small #2}}
\newcommandx{\issue}[2][1=]{\todo[linecolor=red,backgroundcolor=red!25,bordercolor=red,#1]{#2}}
\newcommandx{\counterclaim}[2][1=]{\todo[linecolor=blue,backgroundcolor=blue!25,bordercolor=blue,inline,caption={},#1]{#2}}
\newcommandx{\exactcopy}[2][1=]{\todo[linecolor=darkorange,backgroundcolor=darkorange!25,bordercolor=darkorange!50,inline,#1,noprepend,caption={}]{\small #2}}

\newcommand{\rev}[1]{{#1}} 

\usepackage[pagebackref=true,breaklinks=true,letterpaper=true,colorlinks,bookmarks=false,backref=page]{hyperref}
\usepackage[nameinlink,capitalise,noabbrev]{cleveref}
\crefname{section}{Sec.}{Secs.}
\Crefname{section}{Section}{Sections}
\Crefname{table}{Table}{Tables}
\crefname{table}{Tab.}{Tabs.}

\newcommand{\exhref}[3][blue]{\href{#2}{\color{#1}{#3}}}%

\begin{document}
\newcommand{\project}{\textbf{Beyond Document Page Classification}: Design, Datasets, and Challenges} 

\title{\project}

\newcommand{\superaffil}[2]{\textsuperscript{#1}\,#2}

\author{
\small Jordy Van Landeghem\superaffil{1,2}\and
\small Sanket Biswas\superaffil{3}\and
\small Matthew Blaschko\superaffil{1}\and
\small Marie-Francine Moens\superaffil{1}\and
\and
\footnotesize{ 
  \textsuperscript{1}KU Leuven
  \quad
  \textsuperscript{2}Contract.fit
  \quad
  \textsuperscript{3}Computer Vision Center, Universitat Autònoma de Barcelona}
}




\maketitle

\begin{abstract}
\jordyskeleton{
    This paper highlights the need to bring document classification benchmarking closer to real-world applications, both in the nature of data tested ($X$: multi-channel, multi-paged, multi-industry; $Y$: class distributions and label set variety) and in classification tasks considered ($f$: multi-page document, page stream, and document bundle classification, ...).
    We identify the lack of public multi-page document classification datasets, formalize different classification tasks arising in application scenarios, and motivate the value of targeting efficient multi-page document representations.
    An experimental study on proposed multi-page document classification datasets demonstrates that current benchmarks have become irrelevant and need to be updated to evaluate \textit{complete} documents, as they naturally occur in practice. This reality check also calls for more mature evaluation methodologies, covering calibration evaluation, inference complexity (time-memory), and a range of realistic distribution shifts (e.g., born-digital vs. scanning noise, shifting page order). Our study ends on a hopeful note by recommending concrete avenues for future improvements.}
\end{abstract}

\section{Introduction}

Visual Document Understanding (\DU) comprises a large set of skills, including the ability to holistically process both textual and visual components structured according to rich semantic layouts.
The majority of efforts are directed toward the \rev{application}-directed tasks of classification and extraction of key information (KIE) in visually-rich documents (VRDs).
\textbf{Document classification} (\DC) is a fundamental step in any industrial \DU{} pipeline as it assigns a semantically meaningful category,  routes a document for further processing (towards KIE, fraud checking), or flags incomplete (e.g., missing scans) or irrelevant documents (e.g., \exhref{https://www.family-action.org.uk/content/uploads/2019/07/meals-more-recipes.pdf}{recipe cookbook} in a loan application). 

Documents are intrinsically multi-paged, explaining (partly) why PDF is one of the most popular universal document file formats.\footnote{PDF is the 2nd most popular file format on the web (after HTML and XHTML) following detected MIME types in \href{https://commoncrawl.github.io/cc-crawl-statistics/plots/mimetypes}{CommonCrawl}.}
While \DC{} in information management workflows typically involves multi-page VRDs, current public datasets~\cite{harley2015evaluation, kumar2014structural} only support single-page images and constitute too simplified benchmarks for evaluating fundamental progress in \DC{}.



With the advent of deep learning, the \DU{} field has shifted from region-based analysis to whole-page image analysis.
This shift led to substantial improvements in processing document images with more complex layout variability, exposing the limitations of template-based methods.
Our work highlights the opportunity and necessity of moving \textit{beyond the page} limits toward evaluation on \textit{complete} document inputs, as they prevalently occur (multi-page documents, bundles, page streams, and splits) across various practical scenarios within real-world \DC{} applications, demonstrated in \Cref{fig:heroimage} (see \textit{first page}).




The practical task of long document classification \cite{pramanik2020towards} is largely underexplored due to challenges in computation and how to efficiently represent large multi-modal inputs. 
Additionally, the proximity to applications involves a larger community for conducting research, yet innovations may happen in isolation or are kept back as intellectual property, lacking evaluation on public benchmarks~\cite{gordo2010bag, gordo2013document}, consequently hindering reproducibility and fair comparisons. 


Existing \DC{} methodology is limited to single-page images, and independently and identically distributed (\textit{iid}) settings. We propose an improved methodology that extends its scope to multi-page images and non-iid settings. We also reflect on evaluation practices and put forward more mature evaluation protocols. To better capture the complexity of real-world document handling, we align \DC{} benchmarking closer to practical applications and task formulations.








\noindent Our key contributions can be summarized as follows: 
\begin{itemize}[noitemsep,topsep=-1pt]
  \setlength{\parskip}{2pt}
  \setlength{\itemsep}{0pt plus 0pt}
  \item We have redesigned and formalized multi-page \DC{} scenarios to align fragmented definitions and practices.
  \item We construct and share two novel datasets \texttt{RVL-CDIP\_MP}\footnote{\url{huggingface.co/datasets/bdpc/rvl_cdip_mp}} and \texttt{RVL-CDIP-N\_MP}\footnote{\url{huggingface.co/datasets/bdpc/rvl_cdip_n_mp}} to the community for evaluating multi-page \DC.
  \item We conduct a comprehensive analysis of the novel datasets with different experimental strategies, observing the promise from best-case analysis (+6\% absolute accuracy) from targeting multi-page document representations and inference.   
  \item We overview challenges stalling \DC{} progress, giving concrete guidelines to improve and increase dataset construction efforts. 
\end{itemize}\par
\vspace{1em}

\rev{\section{Problem Formulation}\label{sec:prelim}}
We propose to use formal definitions to better align \DC{} research with real-world document distributions and practices. This will help to standardize \DC{} practices and make it easier to compare different methods.
\noindent
\newline
\jordyskeleton{
Let $\mathcal{X}$ 
denote a space of documents, and let $\mathcal{Y}$ denote the output space as a finite set of discrete labels. 
Document page classification is a prototypical instance of classification \cite{vapnik1992principles}, where the goal is to learn an estimator $f: \mathcal{X} \to \mathcal{Y}$ using $N$ supervised input-output pairs $(X,Y) \in \mathcal{X} \times \mathcal{Y}$ drawn \textit{iid} from an unknown \rev{joint} distribution $P(X,Y)$.}

A \textbf{page} $p$ is a natural classification input that consists of an image $\boldsymbol{v} \in \mathbb{R}^{C \times H \times W}$ (number of channels, height, and width, respectively) with $T$ word tokens $\left\{t_i\right\}_{i=1}^T$ organized according to a layout structure $\left\{\left(x_i^1, y_i^1, x_i^2, y_i^2\right)\right\}_{i=1}^T$, typically referred to as bounding boxes, either coming from Optical Character Recognition (OCR) or natively encoded.

\jordyskeleton{
Note that in practical business settings, VRDs are presented at inference time to a production \DU{} system in different forms: 
    \begin{enumerate}[label=\Roman*.,leftmargin=2\parindent,itemsep=-1mm]
        \item Single page (often scanned or photographed) 
        \item Single document 
        \item Multiple documents 
        \item Multiple pages (often bulk-scanned to a single PDF)
        \item \rev{Single image with multiple localized pages } 
    \end{enumerate}
}

\noindent \rev{\textbf{Classification tasks} \quad} \rev{In a unification attempt, we formalize the different classification inputs and tasks that arise in practical scenarios, as visualized in \cref{fig:heroimage}.} 

\rev{\begin{mydefinition}[Page Classification] \normalfont
(I) A page (as defined above) is categorized with a single category. When only considering the visual modality, the literature refers to it as `document image classification' \cite{harley2015evaluation}.
An estimator for page classification with the input dimensionality ($\mathcal{X}_p$) relative to a page (viz., number of channels, height, and width) is defined as:
\begin{equation}
\begin{gathered}
    f_p: \mathcal{X}_p \to \mathcal{Y},\\ \text{ where }\mathcal{Y} = [C]\text{ for $C$ mutually exclusive categories.}
\end{gathered}
\end{equation}
\end{mydefinition}}

\rev{\begin{mydefinition}[Document Classification] \normalfont
(II) A \textbf{document} $d$ contains a fixed number of $L \in [1,\infty)$ pages, which do not necessarily have the same dimensions (height and width). Albeit a design choice, the input dimensionality is normalized across pages (e.g., $3 \times 224 \times 224$).
Assuming a fixed input dimensionality $(\mathcal{X}_d)$ relative to a document ($L \times C \times H \times W$), a document classifier is defined as:
\begin{equation}
	\begin{gathered}
    f_d: \mathcal{X}_d \to \mathcal{Y}, \\ \text{ where }\mathcal{Y} = [K]\text{ for $K$ mutually exclusive categories.}
\end{gathered}
\end{equation}
\end{mydefinition}}

\rev{Note also the difference in label space between the two previous classification tasks, which can have some overlap for document types that are uniquely identifiable from a single page (e.g., \exhref{https://cartraveldocs.com/wpinstall/wp-content/uploads/2021/01/European-Accident-Statement-details.jpg}{an accident statement form}).}

\rev{\begin{mydefinition}[Document Bundle Classification] \normalfont
(III) A bundle $b$ can contain a variable number of $B$ documents, each with a potentially different amount of $L$ pages. A bundle classifier models a sequence classification problem over multiple documents:
\begin{equation}
	\begin{gathered}
    f_b: \mathcal{X}_b \to \mathcal{Y}\text{, where }\mathcal{Y} \text{ is a product space of $B$ documents, } \\ \mathcal{Y} = \mathcal{Y}_1 \times ... \times  \mathcal{Y}_B \text{, with } \{ \mathcal{Y}_j = [K]: j \in [B]\}.
\end{gathered}
\end{equation}
\end{mydefinition}}

\rev{\begin{mydefinition}[Document Stream Classification] \normalfont
(IV) A page stream $s$ is similar to a document in terms of input (number of pages $L$), albeit typically more varied in content and page formats. 
Page streams can implicitly contain many \rev{different} documents, with pages not necessarily contiguous or even in the right order, as illustrated in \cref{fig:heroimage}.
\begin{multline}
    f_s: \mathcal{X}_d \to \mathcal{Y}\text{, where }\mathcal{Y} \text{ is a product space of $L$ pages, } \\ \mathcal{Y} = \mathcal{Y}_1 \times ... \times  \mathcal{Y}_L \text{, with } \{ \mathcal{Y}_j = [C]: j \in [L]\}. 
\end{multline}
\end{mydefinition}}

A very concrete example of how the label sets $[C]$ and $[K]$ can differ is in a \rev{loan application} use-case where national registry proofs need to be sent: If two pages are sent with the front and back of the \exhref{https://d3i71xaburhd42.cloudfront.net/b3c3afa2e9b13d934a79b4fbe2759ee431b8e77b/1-Figure1-1.png}{ID-card}, $f_s$ requires two labels (\textit{id\_front, id\_back}), whereas $f_d$ requires a single document label (\textit{id\_card}).

\jordyskeleton{A critical note is due to differentiate page stream segmentation (PSS) \cite{gallo2016deep,wiedemann2021multi,mungmeeprued2022tab} and page stream classification as defined above ($f_s$). PSS treats a page stream as a binary classification task to identify document boundaries, without classifying the identified documents afterward. $f_s$ considers the task in one stage where $C$ is constructed in a way to send atomic units such as a \exhref{https://payrollhero.ph/ph/img/product-payslip.jpg}{wage slip} in \Cref{fig:heroimage} for individual downstream processing or it can be combined to a single document label from $[K]$ based on assigned page labels. Two-stage processing is possible by applying PSS as an instance of a $f_s$ classifier with $[C]=\{0,1\}$ where 1 indicates a document boundary, followed by $f_d$.}

\rev{\begin{mydefinition}[Page Splitting] \normalfont
(V) A multi-page image $m$ contains multiple page objects of similar types which can have multiple orientations, page dimensions, and often physical overlap from poor scanning \cite{garimella2016identification}. A standard example involves multiple receipts to be analyzed for reclaiming VAT. While a complete approach will consist of localizing pages (using edge/corner detection, object detection, or instance segmentation) and identifying page types, we will only focus on the latter.
For instance, multi-page splitting can be defined as a preliminary check on how many page types are present in a multi-page image (with input dimensionality similar to a single page $p$):  
\begin{gather}
    f_m: \mathcal{X}_p \to \mathcal{Y}\text{, where }\mathcal{Y} = \mathbb{Z}^{C}.
\end{gather}
\end{mydefinition}}

Payment proofs such as tickets and receipts more often are packed together due to their compactly printed sizes, which would require splitting the unique documents from within a page to send individually for further processing. 
Following the national registry example. another rare yet ``economical" variation for $f_d$ occurs when a single page contains both the front and back of the ID card stitched together. 
These edge cases (rightmost example in \Cref{fig:heroimage}) should be dealt with on a case-by-case basis for how to set up $[K]$ (e.g., specific label: \exhref{https://www.pugetsound.edu/sites/default/files/inline-images/8088_scannedReceiptsExample_0.jpg}{multi-tickets}). 

\rev{The formalisms defined above establishes a taxonomy of \DC{} tasks, which will be retaken in the discussion of challenges to align \DC{} research and applications (\Cref{sec:challenges}).}

\section{Balancing Research \& Applications}

Having established a taxonomy, we further sketch the role of \DC{} in the larger scope of \DU{}, both in the applications and research context. 
We point to related \DU{} benchmarks and describe current \DC{} datasets with their relevant (or missing) properties using the task formalizations.  
Next, we link to related initiatives in dataset construction and calls for reflection on DU practices. 
Finally, we introduce the curated \DC{} datasets to support multi-page \DC{} ($f_d$) benchmarking, which will be used in a further experimental study.


\paragraph{General Benchmarking in VDU:}
In any \textit{industrial application context}  where information transfer and inbound communication services are an important part of the day-to-day processes, a vast number of documents have to be processed. To provide customers with the expected service levels (in terms of speed, convenience, and correctness) a lot of time and resources are spent on categorizing these documents and extracting crucial information. Complex business use cases (such as consumer lending, insurance claims, real estate purchases, and expenditure) involve processing bundles of different documents that clients send via any communication channel. 
For example, obtaining a loan typically entails sending the following documents to prove solvency: a number of \exhref{https://www.forbes.com/advisor/wp-content/uploads/2022/10/image1-7.png}{monthly pay stubs}, \exhref{https://prodblobcdn.azureedge.net/wp/webp/novelty-bank-statement.webp}{bank statements}, \exhref{https://images.sampletemplates.com/wp-content/uploads/2016/10/20144630/Income-Tax-Form-Sample.jpg}{tax forms}, and \exhref{https://www.tradingstandards.uk/media/images/news--policy/press-office/yoticitizencard.jpg?width=390}{national registry proofs}. Furthermore, not all documents are born-digital (BD), and as an artifact of the communication channel (bulk scans/photographs, digitization of physical mail), a single client communication can contain an arbitrary amount of document page images in an unknown order, requiring an $f_s$ classifier. 
\Cref{fig:heroimage} provides an overview of the different \DC{} tasks that arise in application scenarios, which are scarcely covered by \DC{} research benchmarks (see \Cref{tab:dc}).
As RVL-CDIP is the only large-scale non-synthetic \DC{} benchmark, we discuss it in more detail, other dataset descriptions can be found in Supplementary. 

\begin{table*}[ht]
\centering
\begin{tabular}{@{}lllllcc@{}}

\toprule
Dataset      & Size & Data Source                 & Domain              & Task     & OCR & Layout \\ \midrule
IIT-CDIP~\cite{lewis2006building}     &  35.5M    & UCSF-IDL                    & Industry        & Pretrain       &  \xmark   & \xmark       \\
RVL-CDIP~\cite{harley2015evaluation}     &  400K    & UCSF-IDL                    & Industry        & DC     & \xmark    & \xmark     \\
RVL-CDIP-N~\cite{larson2022evaluating}     &  1K    & Document Cloud                    & Industry        & DC     & \xmark    & \xmark     \\
TAB~\cite{mungmeeprued2022tab}     &  44.8K    & UCSF-IDL                    & Industry        & DC     & \xmark    & \xmark     \\
FUNSD~\cite{jaume2019funsd}        &  199    & UCSF-IDL                    & Industry        & KIE      & \cmark    & \xmark       \\
SP-DocVQA~\cite{mathew2020document}       &  12K    & UCSF-IDL                    & Industry        & QA       & \cmark    & \xmark       \\
OCR-IDL~\cite{biten2022ocr}      &  26M    & UCSF-IDL                    & Industry        & Pretrain &  \cmark   & \xmark       \\
FinTabNet~\cite{zheng2021global}    &  89.7K    & Annual Reports S\&P         & Finance             & TSR      &  \xmark   &   \cmark     \\
Kleister-NDA~\cite{kleisterStanislawekGWLK21} & 3.2K     & EDGAR                       & US NDAs             & KIE         & \cmark    & \xmark       \\
Kleister-Charity~\cite{kleisterStanislawekGWLK21} & 61.6K     &  UK Charity Commission      & Legal             &   KIE       & \cmark    & \xmark       \\
DeepForm~\cite{straydeepform}    &  20K    & FCC Inspection & Forms broadcast     &  KIE        & \cmark    & \xmark       \\
TAT-QA~\cite{Zhu_2022}       &   2.8K   & Open WorldBank                            & Finance             & QA       & \cmark    &  \xmark      \\
PubLayNet~\cite{zhong2019publaynet}   & 360K     & PubMed Central              & Scientific  & DLA      & \xmark    & \cmark       \\
DocBank~\cite{li2020docbank}      & 500K     & arxiv                       & Scientific  & DLA         & \cmark    &  \cmark      \\
PubTabNet~\cite{zhong2020image}    &  568K    & PubMed Central              & Scientific  & TSR      & \xmark    & \cmark       \\
DUDE~\cite{vanlandeghem2023document}         &  40K    & Mixed             &  Multi-domain                   &       QA   & \cmark    &  \xmark      \\
Docile~\cite{simsa2023docile}      &   106K   &     EDGAR \& synthetic                        &           Industry          &       KIE   & \cmark    & \xmark       \\ 
CC-PDF~\cite{turski2023ccpdf}         & 1.1M   & Common-Crawl (2010-22)          &  Multi-domain                   &       Pretrain   & \xmark    &  \xmark      \\
\bottomrule
\end{tabular}
\caption{\textbf{DU Benchmarks} with their significant data sources and properties. Acronyms for tasks DC: Document Classification DLA: Document Layout Analysis KIE: Key Information Extraction QA: Question Answering TSR: Table Structure Recognition}
\label{tab:sources}
\end{table*}

\begin{table*}[h]
	    \centering 
    \scalebox{0.9}{
    \makebox[\textwidth]{
    \centering 
    \begin{tabular}{@{}ccccccc@{}}
        \toprule
        \textbf{Dataset} &
        \textbf{Purpose} &
        $\mathbf{\#d}$ &
		$\mathbf{\#p}$ &
		$\mathbf{|\mathcal{Y}|}$ &
        \textbf{Language} &
        \textbf{Color depth} \\ \midrule
    NIST~\cite{dimmick1992nist}             & $f_s$                            &     & 5590                  & 20 & English       & Grayscale \\
    MARG~\cite{long2005image}            & $f_s$                            &     & 1553                  & 2 & English       & RGB \\
        Tobacco-800~\cite{zhu2007automatic}             & $f_s$                            &     & 800                  & 2 & English       & Grayscale \\
         TAB~\cite{mungmeeprued2022tab}             & $f_s$                            &     & 44.8K                  & 2 & English       & Grayscale \\
  Tobacco-3482~\cite{kumar2013unsupervised}             & $f_p$                            &     & 3482                  & 10 & English       & Grayscale \\
        RVL-CDIP~\cite{harley2015evaluation}                 & pre-training, $f_p$ &     & 400K                  & 16 & English       & Grayscale \\
        RVL-CDIP-N~\cite{larson2022evaluating}               & $f_p$, OOD                       &     & 1002                  & 16 & English       & RGB       \\
        RVL-CDIP-O~\cite{larson2022evaluating}               & $f_p$, OOD                       &     & 3415                  & 1  & English/Mixed & RGB       \\
        \texttt{RVL-CDIP\_MP}    & $f_d$                                  & $\pm$400K & $\mathbb{E}[L]=5$ & 16 & English       & Grayscale \\
        \texttt{RVL-CDIP-N\_MP} & $f_d$, OOD                             & 1002   & $\mathbb{E}[L]=10$ & 16 & English       & RGB       \\
        \bottomrule
    \end{tabular}
}}
 \caption{\textbf{Statistical Comparison} of public and proposed extended multi-page \DC{} datasets. OOD refers to out-of-distribution detection. $\#d$ and $\#p$ refer to number of documents or pages, respectively. For the novel MP datasets, we report the average number of pages. }
\label{tab:dc}
\end{table*}

Current state-of-the-art DU research based approaches~\cite{appalaraju2021docformer,huang2022layoutlmv3,li2022dit} leverage the ``pre-train and fine-tune" procedure that performs significantly well on popular DU benchmarks~\cite{harley2015evaluation,huang2019icdar2019,zhong2019publaynet,jaume2019funsd} (see \Cref{tab:sources}). 
 However, their performance drops significantly when exposed to real-world business use cases mainly due to the following reasons: 
 (1) The models are limited to modeling page-level context due to heavy compute requirements (e.g., quadratic complexity of self-attention \cite{vaswani2017attention}), effectively treating each document page as conditionally independent and potentially missing out on essential classification cues. (2) The methods are heavily reliant on the quality of OCR engines to extract spatial local information (i.e. mostly at word level) suitable to solve downstream benchmark tasks; but fail to \textit{generalize} well on business documents. (3) Existing datasets used for pre-training \cite{lewis2006building,harley2015evaluation} are different in terms of domain, content, and visual appearance from many downstream \DC{} tasks (detailed in \Cref{sec:X}).
 Therefore, it can be challenging for industry practitioners to choose a specific model to fine-tune for the \DC{} use cases and task specifics that they commonly encounter.

\noindent \textbf{RVL-CDIP} The Ryerson Vision Lab Complex Document Information Processing \cite{harley2015evaluation} dataset used the original IIT-CDIP (The Illinois Institute of Technology dataset for Complex Document Information Processing) \cite{lewis2006building} metadata to create a new dataset for document classification. 
It was created as the equivalent of ImageNet in the \DU{} field, which invited a lot of multi-community (Computer Vision, NLP) efforts to solve this dataset. It consists of low-resolution, scanned documents belonging to one of 16 classes such as \textit{letter, form, email, invoice}. 

\paragraph{Proposed Datasets}\label{novel-datasets}

\jordyskeleton{\noindent\texttt{RVL-CDIP\_MP} is our first contribution to retrieve the original documents of the IIT-CDIP test collection which were used to create RVL-CDIP. Some PDFs or encoded images were corrupt, which explains that we have around 500 fewer instances. By leveraging metadata from OCR-IDL \cite{biten2022ocr}, we matched the original identifiers from IIT-CDIP and retrieved them from IDL using a conversion. 
However, the same caveats for RVL-CDIP apply.} 

\noindent\texttt{RVL-CDIP\_MP-N} can serve its original goal as a covariate shift test set, now for multi-page document classification. We were able to retrieve the original full documents from DocumentCloud and Web Search.
As no existing large-scale datasets include granular page-level labeling (in terms of $[C]$) for multi-page documents, we could not create a benchmark for evaluating $f_s$.
\Cref{app:viz} points to visualizations from the proposed datasets.

\paragraph{Related Initiatives} \label{sec:relwork}
 General benchmarking challenges have driven the \DU{} research community to set the seed for initiatives to create its own document-oriented “ImageNet”  \cite{russakovsky2015imagenet} challenge, over which multiple long-term grand challenges can be defined (\href{http://cvit.iiit.ac.in/deepdoc2022/}{deepdoc2022}, 
\href{http://cvit.iiit.ac.in/scaldoc2023/}{scaldoc2023}). 
In another task paradigm, DocuVQA, there have been efforts in the same spirit to redirect focus to multi-page documents \cite{tito2022hierarchical,dude2023icdar}. For the task of KIE, \cite{rossum2022practicalbenchmarks} launched a similar call for practical document benchmarks closer to real-world applications.
While these initiatives demonstrate a similar-looking future direction, our contribution goes beyond introducing novel datasets and seeks to guide the complete methodology of \DC{} benchmarking.

        



\section{Experimental Study}\label{sec:exp}


\jordyskeleton{To classify a multi-page document, one might ask the question ``\textit{Why not just predict based on the first page? What would be the gain of processing all pages? What baseline inference strategies can be applied to classify a multi-page document?}". This prompted us to put these assumptions to the test in a small motivating study\footnote{Code provided at: \url{https://huggingface.co/bdpc/src}}.

As current public datasets only support page classification, we have extended some existing \DC{} datasets to already enable testing a slightly more realistic, yet more complex document classification scenario ($f_d$).}

We have reconstructed the original PDF data of the \DC{} datasets in \Cref{novel-datasets}. 
The goal of this experiment is to tease some issues and strategies when naively scaling beyond page-level \DC.
Our baseline of choice is the document foundation model DiT-Base \cite{li2022dit}, which as a visual-only $f_p$ is competitive with more compute-intensive multimodal, OCR-based pipelines \cite{huang2022layoutlmv3,appalaraju2021docformer,tang2023unifying}. 

\begin{table}[h]
    \centerfloat
 \centering
        \resizebox{0.7\columnwidth}{!}{
    \begin{tabular}{@{}ccc@{}}
        \toprule
        \textbf{Inference} & \multicolumn{1}{c}{\textbf{Strategy}} & \multicolumn{1}{c}{\textbf{Scope}} \\ \midrule
        \textit{sample}      & first           & page                           \\
        & second          & page                           \\
        & last            & page                            \\
        \textit{sequence}    & max confidence & page  \\
        \multicolumn{1}{l}{} & soft voting    & page  \\
        \multicolumn{1}{l}{} & hard voting    & page  \\ 
        \textit{grid}        & grid            & document \\
        \textit{document}    & \tiny{(not tested)} & document \\
        \bottomrule
    \end{tabular}}
    \caption{\textbf{Tested inference methods} to classify multi-paged documents and simulate a true document classifier $f_d$. Scope refers to the independence assumption taken at inference time.}
    \label{tab:infstrats}
\end{table}

\jordyskeleton{\Cref{tab:infstrats} overviews some straightforward inference strategies. 
Consider the simplest inference strategy is to \textit{sample} a given page with index $l \in [L]$ (or in our case $\{1,2,L-1\}$) from $\hat{y}^l=[f_p(x)]^l$. The \textit{sequence} strategies mainly differ in how the final prediction $\hat{y}$ is obtained from predictions per page, assuming a probabilistic classifier $\tilde{f_p}: \mathcal{X}_p \to [0,1]^K$.

\begin{equation}
\text{MaxConf}(x,y) = \argmax_{\substack{l \in [L] \\ k \in [K]}} [\tilde{f_p}(x,y)]_{k}^l  
\end{equation}
\begin{equation}
\text{SoftConf}(x,y) = \argmax_{k \in [K]} \sum_{l=1}^{L} [\tilde{f_p}(x,y)]^l  
\end{equation}
\begin{equation}
\text{HardVote}(x,y) = \argmax_{k \in [K]} \sum_{l=1}^{L} e_{\hat{y}^l},
\end{equation} with $e$ a one-hot vector of size $K$. 
The \textit{grid} strategy is intuitive as we tile all page images in an equal-sized grid that trades off the resolution to jointly consume all document pages. While results in this experiment with fairly low grid resolution (224 x 224) are poor, variations (with aspect-preserving \cite{lee2022pix2struct} or layout density-based scaling) deserve to be further explored.}

\begin{table}[h]
\centering
\npdecimalsign{.}
\nprounddigits{3}
\begin{tabular}{|l|n{1}{3}n{1}{3}n{1}{3}|n{1}{3}n{1}{3}|}
\toprule
\textbf{Strategy} & \text{Acc}$\uparrow$ & \text{F1}$\uparrow$ & \text{F1}$_{M}\uparrow$ & \text{ECE}$\downarrow$ & \text{AURC}$\downarrow$ \\
\midrule
$f_p\$$ \cite{li2022dit} & 
93.34483362084052
&  93.35052090923793 & 93.33492762947787  &  0.07530557978075443 &  0.010363491269748872 \\
\hline\hline
first          &  91.291066 &  91.286111 &  91.271228 &  0.073018 &   0.014433 \\
second         &  87.294671 &  87.304522 &  87.276629 &  {\npboldmath} 0.070183 &  0.029450 \\
last           &  85.091146 &  85.059711 &  85.028274 &  0.072225 &  0.037787 \\
MaxConf &  {\npboldmath} 91.407463 &  {\npboldmath} 91.452536 &  {\npboldmath} 91.343696 &  0.123796 &  0.005873 \\
SoftVote    &  91.219634 &  91.185159 &  91.235840 &  0.134472 &  {\npboldmath} 0.004123 \\
HardVote    &  85.995492 &  86.182321 &  85.780544 &  0.085059 &  0.018197 \\
grid           &  72.642124 &  72.044572 &  73.266158 &  0.108862 &  0.041505 \\
\bottomrule
    \end{tabular}
    \caption{Base classification accuracy of DiT-base \cite{li2022dit} (finetuned on RVL-CDIP) evaluated on the test set of \texttt{RVL-CDIP\_MP} per baseline $f_d$ strategy. Best results per metric are boldfaced. \$ refers to our reproduction of results.} 
    \label{tab:res}
\end{table}

\begin{table}[h]
\centering
\npdecimalsign{.}
\nprounddigits{3}
\begin{tabular}{|l|n{1}{3}n{1}{3}n{1}{3}n{1}{3}n{1}{3}|}
\toprule
\textbf{Strategy} & \text{Acc}$\uparrow$ & \text{F1}$\uparrow$ & \text{F1}$_{M}\uparrow$ & \text{ECE}$\downarrow$ & \text{AURC}$\downarrow$ \\
\midrule
$f_p$ \cite{larson2022evaluating} & 78.64271457085829  &  81.94724973834633 & 60.56424622312553  & 0.10465847378124496  &  0.07631596919925768 \\
\hline\hline
    first          &  {\npboldmath}78.760163 &  {\npboldmath}75.316412 &  {\npboldmath}60.801032 &  0.143747 &  {\npboldmath}0.024576 \\
    second         &  64.939024 &  58.740976 &  50.773314 &  0.131963 &  0.070656 \\
    last           &  64.227642 &  58.192229 &  48.859496 &  0.128158 &  0.073811 \\
    MaxConf &  76.321138 &  72.855020 &  57.469959 &  0.179588 &  0.042315 \\
    SoftVote    &  73.983740 &  69.162888 &  56.485704 &  0.183370 &  0.039250 \\
    HardVote    &  67.479675 &  63.187572 &  52.234591 &  0.110275 &  0.087816 \\
    grid           &  47.755102 &  40.644680 &  38.584193 &  {\npboldmath}0.101567 &  0.169908 \\
    \bottomrule
    \end{tabular}
    \caption{Base classification accuracy of DiT-base \cite{li2022dit} (finetuned on RVL-CDIP) evaluated on the test set of \texttt{RVL-CDIP\_N\_MP} per baseline $f_d$ strategy. Best results per metric are boldfaced. } 
    \label{tab:res-N}
\end{table}

Following similar calls (discussed infra. \Cref{sec:eval}) in the \DU{} literature \cite{vanlandeghem2023document} to establish calibration and confidence ranking as default evaluation metrics, we include Expected Calibration Error (\ECE) \cite{niculescu2005predicting,naeini2015obtaining,guo2017calibration} to evaluate top-1 prediction miscalibration and Area-Under-Risk-Coverage-Curve (\AURC) \cite{geifman2017selective,jaeger2023a} to measure selective (proportion of test set\%) accuracy. 
    
    Results in \Cref{tab:res,tab:res-N} demonstrate that classifying by only the first page is a solid strategy, with performance dropping when considering only later pages. 
    Maximum confidence and soft voting require $L$ (pages) times more processing, yet attain similar performance as the best single-page prediction.
    However, this could be attributed to two factors: i) dataset creation bias since \cite{harley2015evaluation} constructed RVL-CDIP from a page of each original .tiff file, for which the label was kept if it belonged to one of the 16 categories, whereas RVL-CDIP-N \cite{larson2022evaluating} consistently chose the first-page; ii) documents are fashioned in a summary-detail or top-down content structure over pages. To confirm the validity of the latter hypothesis, more robust experiments on more fine-grained labeled \DC{} are needed. 
    

    The results from \Cref{tab:res} and \Cref{tab:res-N} can be interpreted as an upper bound (iid) and a loose lower bound (non-iid, yet related), respectively.
    For the former, MaxConf is the most accurate, yet compared to SoftVote has worse AURC, potentially making SoftVote a better candidate for industry use where controlled risk is more valued.
    While this trend is not reproduced in \texttt{RVL-CDIP\_N\_MP}, it can be explained by the more consistent first-page labeling, adding distracting classification cues from later pages.  
    
\begin{table}[h]
\centering
\npdecimalsign{.}
\nprounddigits{3}
\begin{tabular}{|l|l|n{1}{3}n{1}{3}|}
\toprule
\textbf{Dataset} & \textbf{Strategy} & \text{Acc}$\uparrow$ & $\Delta$ \\
\midrule

        {\small \texttt{RVL-CDIP\_MP}} & first+second$^{(*)}$                          & 93.7950721153846                      & 2.5040061153845983 \\
        & first+last$^{(*)}$                            & 93.6748798076923                      & 2.383813807692306 \\
       & second+last$^{(*)}$                           & 89.7085336538462                      & -1.5825323461537977 \\
         & first+second/last$^{(*)}$                     & {\npboldmath} 94.454              & 3.162933999999993 \\ \bottomrule

        {\small \texttt{RVL-CDIP\_N\_MP}} & first+second$^{(*)}$                          & 83.63821138211383                      & 4.878048382113818 \\
        & first+last$^{(*)}$                            &         83.13008130081301             & 4.369918300812998 \\
       & second+last$^{(*)}$                           & 71.54471544715447                     & -7.215447552845532 \\
         & first+second/last$^{(*)}$                     & {\npboldmath} 84.55284552845529              & 5.792682528455288 \\ \bottomrule



    \bottomrule
    \end{tabular}
    \caption{Best-case classification accuracy indicated with $^{(*)}$ when combining 'knowledge' over different pages. $\Delta$ refers to the absolute difference with the first page only.} 
    \label{tab:res-best}
\end{table}


    To answer what can be gained from processing a multi-page document in a single shot, \Cref{tab:res-best} reports a best-case error analysis, where a page prediction is counted as correct if the model would have had access to the other pages. This is calculated by using a bit-wise OR operation between the one-hot vectors $(\mathbb{I}[y==\hat{y}])$ expressing correctness for each strategy model. As a proof of concept, this shows that targeting multi-page document representations and inference is a promising avenue to improve \DC{}.

\section{Challenges and Guidelines}\label{sec:challenges}

Following \rev{the introduced task formalizations of \Cref{sec:prelim}}, we claim that the distribution on which document classification is currently evaluated publicly and the real-world distributions have heavily diverged.
Additionally, our experimental validation on the novel datasets demonstrated the potential of multi-page \DC, empirically reinforcing our call to action on improving \DC{} methodologies.
Let $P^A(X,Y)$ and $P^R(X,Y)$ denote those two distinct distributions, \textit{real-world applications} and \textit{research} respectively. Further, we will characterize the specific divergences with concrete examples and suggestions for better alignment.

    \subsection{Divergence of tasks: $f$}\label{sec:tasks} 

The challenge of directly processing multi-paged documents is typically avoided by current \DC{} models which only support single-page images \cite{Powalski2021GoingFB,appalaraju2021docformer,huang2022layoutlmv3,gu2021unidoc,li2021selfdoc,kim2021donut,tang2023unifying,lee2022pix2struct}.
Whenever a new DU model innovation happens, the impact for document classification is publicly only measured on the first task scenario (e.g., $f_p$ on RVL-CDIP), whereas production DU systems more often need to deal with the other settings (II,III,IV,V) \rev{in \Cref{fig:heroimage}}.
Moving beyond the limited page image context will test models' ability to sieve through potentially redundant and noisy signals, as the classification can be dependent on very local cues such as a single title on the first page or the presence of signatures on the last page. Without any datasets to test this ability, we also cannot blindly assume that we can simply scale $f_p$ classifiers to take in more context or that aggregating isolated predictions over single pages is a future-proof (performant and efficient) strategy, as our experiments have shown.

While $p$ is a natural processing unit for humans, acquiring supervised annotations for every single page can be more expensive than attaching a single content-based label (from $[K]$) to a multi-page document. However, fine-grained labeling with $f_s$ could allow for more targeted and constrained KIE, as knowing a certain page $l$ has label $y^l=\text{id\_front} \in [C]$
will allow you to focus on specific entities such as \textit{national registry number, date/place of birth}. Ultimately, these classification task formulations can also help one consider how to set up $f$ directly and annotate document inputs, depending on the \DC{} use-case.

    \subsection{Divergence of label space: $Y$}\label{sec:Y}

Current benchmarks often use simplified label sets that are difficult to reconcile with industry requirements. While RVL-CDIP is the de facto standard for measuring performance on $f_p$ \DC, recent research \cite{larson2023labelnoise} has revealed several undesirable characteristics. 
It supports only 16 labels that pertain to a limited yet generic subset of business documents, which is far from the 1K classes in ImageNet on whose image it was modeled. Real-world \DC{} use cases typically support a richer number of classes ($K \sim$ 50-400). 
RVL-CDIP suffers from substantial label noise, estimated to be higher than current state-the-art $f_p$ error rates (see \cite{larson2023labelnoise} for a detailed analysis) which are overfit to noise. Due to the absence of original labeling guidelines, the labels in RVL-CDIP can be ambiguous, containing disparate subtypes (e.g., business cards in the \textit{resume} category), and inconsistencies between classes (cheques present in both \textit{budget} and \textit{invoice}). Other errors include (near-)duplicates causing substantial overlap between train and test distributions, corrupt documents, and plain mislabeling. However, many common CV benchmarks are plagued by similar issues \cite{beyer2020we} and would benefit from relabeling campaigns \cite{yun2021re} to maintain their relevance.

Considering the above, multi-label classification (not covered explicitly in \Cref{sec:tasks}) could be a solution to resolve label ambiguities, yet this requires absolute consistency in label assignments, which when lacking introduces even more label noise. 
The highest labeling quality could arise from consistent labeling at the page level and hierarchically aggregating page labels ($C \to K$), yet granular annotations are more expensive to obtain. Alternatively, it may be better to follow the mutually exclusive and collectively exhaustive (MECE) principle \cite{chevallier2016strategic} to construct label sets at the document level.

Finally, an overlooked aspect of current benchmarks is that label sets $[K]$ can be constructed based on some business logic, where a very local cue can lead to a class assignment such as some checked box on page 26. Admittedly, this does conflate the tasks of document object detection, KIE, and \DC{} within a single label set. 
However, the current focus on classes with plenty of evidence across a document, with more global classification cues, should be balanced with document types that rely on local cues.

Taking the above issues into account, the community should work together towards developing more effective and realistic \DC{} datasets that better align with the needs of industry practitioners. While tackling the challenge of $Y$ divergence was out-of-scope for the contributed datasets, the next Subsection gives systematic recommendations for obtaining better future \DC{} benchmarks.

    \subsection{Divergence of input data: $X$}\label{sec:X}

We offer suggestions for future benchmark construction efforts such that they take into account what properties are currently unaccounted for, organically improving on our first pursuit towards multi-page \DC{} benchmarking.

We argue that current \DU{} benchmarks fail to account for many real-world document data complexities: multiple pages, the distinction between born-native, (mobile) scanned documents,  accounting for differences in quality, orientation, and resolution.
Additionally, the UCSF Industry Document Library (and in consequence all \DC{} datasets drawn from this source)
contains mostly old (estimated period 1950s to 2002), type-written black and white documents, while in reality, modern documents can have multiple channels, colors, and (embedded) fonts varying in size, typeface, typography. 
Recently, there have been efforts to collect more modern VRD benchmarks for tasks such as DocVQA \cite{mathew2022infographicvqa,vanlandeghem2023document}, KIE \cite{simsa2023docile}, DLA \cite{pfitzmann2022doclaynet}.
Modern VRDs contain visual artifacts such as logos, checkboxes, barcodes, and QR codes; geometric elements such as rectangles, arrows, charts, diagrams, ..., all of which are not frequently encountered with the same variety in current benchmarks.
Future \DC{} benchmarks should incorporate modern VRDs to bring more diversity and variability in input data.

\begin{figure}
    \centering
    \includegraphics[height=3cm, width=8cm]{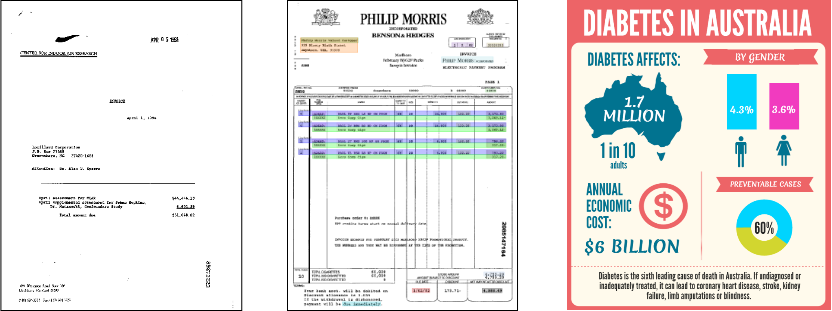}
    \caption{\textbf{Divergence of input data.} The first image is an example from \DC{} benchmark RVL-CDIP~\cite{harley2015evaluation}, the second one from Docile~\cite{simsa2023docile} for KIE, while the third one comes from Info-VQA~\cite{mathew2022infographicvqa}, illustrating the visual-layout richness of modern VRDs vs. the monotonicity of most \DC{} document data.}
    \label{fig:modernity}
\end{figure}

When developing DU models, it is therefore important to consider the role of vision, language, and layout and how these are connected to the classification task. 
For example, current datasets are based on tobacco industry documents containing very domain-specific language, which a less robust classifier can overfit (e.g., the spurious cue of a particular cigarette brand indicates an invoice).
We highlight that document data can be multi-lingual, and code-switching is fairly common in document-based communications. For instance, an email may be in one language while the attachment is in another language.

In summary, future benchmarks must contain multi-page, multi-type, multi-industry (e.g., retail vs. medical invoice)
, multi-lingual documents with a wide range of document data complexities to build and test generic \DC{} systems. 

\jordyskeleton{
The community should explore potential solutions to the lack of adequate datasets for testing \DC{} models such as i) leveraging public document collections, ii) synthetic generation, and iii) anonymization.}

\paragraph{Public document collections:} There are increasingly more (non-profit) organizations (e.g., \href{https://www.documentcloud.org/home}{DocumentCloud}), governments (\href{https://www.sec.gov/edgar/search-and-access}{SEC EDGAR}, financial institutions (\href{https://documents.worldbank.org/en/publication/documents-reports}{World Bank Documents \& Reports}), and charities (\href{https://www.guidestar.org/}{Guidestar}) that make business-related documents publicly available for transparency in their operations and archival/research purposes. These collections provide datasets that are closer to real-world scenarios. However, these documents are typically unlabelled, although annotations could be crowd-sourced through combined funding from interested parties. Since most document data sources restrict automated crawling or document scraping, future dataset constructions will require some cooperation and creativity, whilst fulfilling licensing, ethical, and legal requirements. A specific highlighted initiative is CC-PDF \cite{turski2023ccpdf}, which collected modern, multi-lingual VRDs from CommonCrawl for future use.

\paragraph{Data synthesis:} This alternative was suggested by prior work on KIE \cite{bensch2021key,rossum2022practicalbenchmarks} and DLA~\cite{biswas2021docsynth}for generating business and scientific documents. \cite{simsa2023docile} followed up on this, delivering a large-scale KIE dataset with 6K real documents annotated and 100K synthetic examples. 
However, synthetic generation can be challenging to simulate real-world documents with similar data and classification complexity.

\paragraph{Anonymization} can be a viable option to construct a \DC{} dataset without compromising ethical guidelines and privacy regulations. This process involves removing, masking, replacing, or obfuscating data so that document content can no longer be attributed to an individual or entity. 
For example, one should remove names, addresses, and identifying information such as social security numbers or replace it with a textual tag ({\small\textsc{[social-security-number]}}) or similar pattern (e.g., \href{https://faker.readthedocs.io/en/master/}{Faker}). While this process is not viable for creating KIE datasets, KIE can play a big role in semi-automatically anonymizing documents \cite{glaser2021anonymization,pilan2022text}. 
Companies may be hesitant to make document collections public due to concerns about privacy, confidentiality and GDPR compliance. While anonymization can be an effective method, it should be approached with caution as potential risks of re-identification can make someone with originally good intentions legally liable. 
A potential side-step can be investing in privacy-preserving federated learning (e.g., \href{https://benchmarks.elsa-ai.eu/?ch=2}{PFL-DocVQA}) to allow access to private industry document data.

\subsection{Maturity of evaluation methodology}\label{sec:eval}

\jordyskeleton{
Most \DC{} models are evaluated using predictive performance metrics such as accuracy, precision-recall, and F1-score on \textit{iid} test sets. However, in user-facing applications, calibration can be as important as accuracy \cite{niculescu2005predicting,naeini2015obtaining,guo2017calibration}. Even more so, when the confidence estimation of a \DC{} is used to triage predictions to either an automated flow or manual processing by a human.
Once a \DC{} is in production, the \textit{iid} assumption will start to break, which would recommend a priori testing of robustness against various sources of noise (OCR, subtle template changes, wording or language variations, ...) and expected distribution shifts (born-digital-scanning artifacts, shifting page order, page copies, irrelevant or out-of-scope documents, novel document classes, concept drift, ...).}

\jordyskeleton{Nevertheless, we observe only a few applications in \DC{} (only reported on $f_p$) of more mature evaluation protocols \cite{jaeger2023a} beyond predictive performance. Notable exceptions include covariate shift detection from document image augmentations \cite{maini2022augraphy}, sub-class shift and generalization in \cite[RVL-CDIP-N]{larson2022evaluating}, out-of-distribution detection \cite[RVL-CDIP-O]{larson2022evaluating}, and cross-domain generalization \cite[(RVL-CDIP $\leftrightarrow$ Tobacco-3482)]{bakkali2023vlcdoc}. However, the results on the latter can be misleading as both datasets are drawn from a similar source distribution. Another gap in \DC{} benchmarking concerns evaluating selective classification \cite{geifman2017selective,jaeger2023a}, which is closer to the production value evaluation of how many documents can be automated without any human assistance. 
}

\jordyskeleton{
Another interesting evaluation protocol concerns \textit{out-of-the-box} performance or how data-hungry/sample-efficient a certain model is. In practice, few-shot learning from minimal annotations is a highly valued skill. This few-shot learning evaluation protocol has been applied in \cite{sage2021data} with different data regimes.
Finally, inference complexity (time-memory) has been brought back to the attention of OCR-free models \cite{kim2021donut}, which we believe will be the key to measuring when scaling solutions to multi-page documents.
}

\section{Closing Remarks}


Our work represents a pivotal step forward in establishing multi-page \DC{} by proposing a comprehensive benchmarking and evaluation methodology. Thereby, we have addressed longstanding challenges and limitations \Cref{sec:challenges} that have hindered progress in the field. As motivated in our experimental study, we have proven the need to advance multi-page document representations and inference. 

Following up on this, we provide recommendations for future \DC{} dataset construction efforts pertaining to the type and nature of document data, variety in and quality of the classification label set, with a focus on particular \DC{} scenarios closer to applications, and finally how future progress should be measured.
Nonetheless, we are hopeful that the \DU{} community can come together on these shortcomings and apply the lessons from this reality check. Extending the applicability of current state-of-the-art models in \DU{} to multi-page documents needs further exploration, which will go hand in hand with benchmark creation initiatives or incorporating multiple \DC{} task annotation layers on a single dataset. 

\section*{Acknowledgements}

\noindent
The authors acknowledge the financial support of VLAIO (Flemish Innovation \& Entrepreneurship) through the Baekeland Ph.D. mandate (HBC.2019.2604), and Ph.D. Scholarship from AGAUR (2023 FI-3-00223). Many thanks to Rubèn Pérez Tito and Stefan Larson for guidance on curating the proposed datasets. 

{\small 
\bibliographystyle{ieee_fullname}
\bibliography{main}

\begin{thebibliography}{10}\itemsep=-1pt

\bibitem{appalaraju2021docformer}
Srikar Appalaraju, Bhavan Jasani, Bhargava~Urala Kota, Yusheng Xie, and R
  Manmatha.
\newblock Docformer: End-to-end transformer for document understanding.
\newblock In {\em Proceedings of the IEEE/CVF International Conference on
  Computer Vision}, pages 993--1003, 2021.

\bibitem{bakkali2023vlcdoc}
Souhail Bakkali, Zuheng Ming, Mickael Coustaty, Mar{\c{c}}al Rusi{\~n}ol, and
  Oriol~Ramos Terrades.
\newblock Vlcdoc: Vision-language contrastive pre-training model for
  cross-modal document classification.
\newblock {\em Pattern Recognition}, 139:109419, 2023.

\bibitem{bensch2021key}
Oliver Bensch, Mirela Popa, and Constantin Spille.
\newblock Key information extraction from documents: Evaluation and generator.
\newblock In {\em European Semantic Web Conference (ESWC 2021) and 2nd
  International Workshop, in conjunction with ESWC 2021: Workshop: Deep
  Learning meets Ontologies and Natural Language Processing}, 2021.

\bibitem{beyer2020we}
Lucas Beyer, Olivier~J H{\'e}naff, Alexander Kolesnikov, Xiaohua Zhai, and
  A{\"a}ron van~den Oord.
\newblock Are we done with imagenet?
\newblock {\em arXiv preprint arXiv:2006.07159}, 2020.

\bibitem{biswas2021docsynth}
Sanket Biswas, Pau Riba, Josep Llad{\'o}s, and Umapada Pal.
\newblock Docsynth: a layout guided approach for controllable document image
  synthesis.
\newblock In {\em International Conference on Document Analysis and
  Recognition}, pages 555--568. Springer, 2021.

\bibitem{biten2022ocr}
Ali~Furkan Biten, Ruben Tito, Lluis Gomez, Ernest Valveny, and Dimosthenis
  Karatzas.
\newblock Ocr-idl: Ocr annotations for industry document library dataset.
\newblock {\em arXiv preprint arXiv:2202.12985}, 2022.

\bibitem{chevallier2016strategic}
Arnaud Chevallier.
\newblock {\em Strategic thinking in complex problem solving}.
\newblock Oxford University Press, 2016.

\bibitem{dimmick1992nist}
DL Dimmick, MD Garris, and CL Wilson.
\newblock Nist special database 6. structured forms database 2.
\newblock Technical report, Technical report, National Institute od Standards
  and Technology. Advanced~…, 1992.

\bibitem{gallo2016deep}
Ignazio Gallo, Lucia Noce, Alessandro Zamberletti, and Alessandro Calefati.
\newblock Deep neural networks for page stream segmentation and classification.
\newblock In {\em 2016 International Conference on Digital Image Computing:
  Techniques and Applications (DICTA)}, pages 1--7. IEEE, 2016.

\bibitem{garimella2016identification}
Siddharth Garimella.
\newblock Identification of receipts in a multi-receipt image using spectral
  clustering.
\newblock {\em International Journal of Computer Applications}, 155(2), 2016.

\bibitem{geifman2017selective}
Yonatan Geifman and Ran El-Yaniv.
\newblock Selective classification for deep neural networks.
\newblock {\em Advances in neural information processing systems}, 30, 2017.

\bibitem{glaser2021anonymization}
Ingo Glaser, Tom Schamberger, and Florian Matthes.
\newblock Anonymization of german legal court rulings.
\newblock In {\em Proceedings of the Eighteenth International Conference on
  Artificial Intelligence and Law}, pages 205--209, 2021.

\bibitem{gordo2010bag}
Albert Gordo and Florent Perronnin.
\newblock A bag-of-pages approach to unordered multi-page document
  classification.
\newblock In {\em 2010 20th International Conference on Pattern Recognition},
  pages 1920--1923. IEEE, 2010.

\bibitem{gordo2013document}
Albert Gordo, Mar{\c{c}}al Rusinol, Dimosthenis Karatzas, and Andrew~D
  Bagdanov.
\newblock Document classification and page stream segmentation for digital
  mailroom applications.
\newblock In {\em 2013 12th International Conference on Document Analysis and
  Recognition}, pages 621--625. IEEE, 2013.

\bibitem{gu2021unidoc}
Jiuxiang Gu, Jason Kuen, Vlad~I Morariu, Handong Zhao, Rajiv Jain, Nikolaos
  Barmpalios, Ani Nenkova, and Tong Sun.
\newblock Unidoc: Unified pretraining framework for document understanding.
\newblock {\em Advances in Neural Information Processing Systems}, 34:39--50,
  2021.

\bibitem{guo2017calibration}
Chuan Guo, Geoff Pleiss, Yu Sun, and Kilian~Q. Weinberger.
\newblock On calibration of modern neural networks.
\newblock In {\em Proceedings of the 34th International Conference on Machine
  Learning - Volume 70}, ICML'17, page 1321–1330, 2017.

\bibitem{harley2015evaluation}
Adam~W Harley, Alex Ufkes, and Konstantinos~G Derpanis.
\newblock Evaluation of deep convolutional nets for document image
  classification and retrieval.
\newblock In {\em 2015 13th International Conference on Document Analysis and
  Recognition (ICDAR)}, pages 991--995. IEEE, 2015.

\bibitem{huang2022layoutlmv3}
Yupan Huang, Tengchao Lv, Lei Cui, Yutong Lu, and Furu Wei.
\newblock Layoutlmv3: Pre-training for document ai with unified text and image
  masking.
\newblock {\em arXiv preprint arXiv:2204.08387}, 2022.

\bibitem{huang2019icdar2019}
Zheng Huang, Kai Chen, Jianhua He, Xiang Bai, Dimosthenis Karatzas, Shijian Lu,
  and CV Jawahar.
\newblock Icdar2019 competition on scanned receipt ocr and information
  extraction.
\newblock In {\em 2019 International Conference on Document Analysis and
  Recognition (ICDAR)}, pages 1516--1520. IEEE, 2019.

\bibitem{jaeger2023a}
Paul~F Jaeger, Carsten~Tim L{\"u}th, Lukas Klein, and Till~J. Bungert.
\newblock A call to reflect on evaluation practices for failure detection in
  image classification.
\newblock In {\em International Conference on Learning Representations}, 2023.

\bibitem{jaume2019funsd}
Guillaume Jaume, Hazim~Kemal Ekenel, and Jean-Philippe Thiran.
\newblock Funsd: A dataset for form understanding in noisy scanned documents.
\newblock In {\em 2019 International Conference on Document Analysis and
  Recognition Workshops (ICDARW)}, volume~2, pages 1--6. IEEE, 2019.

\bibitem{kim2021donut}
Geewook Kim, Teakgyu Hong, Moonbin Yim, Jinyoung Park, Jinyeong Yim, Wonseok
  Hwang, Sangdoo Yun, Dongyoon Han, and Seunghyun Park.
\newblock Donut: Document understanding transformer without ocr.
\newblock {\em arXiv preprint arXiv:2111.15664}, 2021.

\bibitem{kumar2013unsupervised}
Jayant Kumar and David Doermann.
\newblock Unsupervised classification of structurally similar document images.
\newblock In {\em 2013 12th International Conference on Document Analysis and
  Recognition}, pages 1225--1229. IEEE, 2013.

\bibitem{kumar2014structural}
Jayant Kumar, Peng Ye, and David Doermann.
\newblock Structural similarity for document image classification and
  retrieval.
\newblock {\em Pattern Recognition Letters}, 43:119--126, 2014.

\bibitem{larson2022evaluating}
Stefan Larson, Gordon Lim, Yutong Ai, David Kuang, and Kevin Leach.
\newblock Evaluating out-of-distribution performance on document image
  classifiers.
\newblock In {\em Thirty-sixth Conference on Neural Information Processing
  Systems Datasets and Benchmarks Track}, 2022.

\bibitem{larson2023labelnoise}
Stefan Larson, Gordon Lim, and Kevin Leach.
\newblock On evaluation of document classification with rvl-cdip.
\newblock In {\em Proceedings of the 17th Conference of the European Chapter of
  the Association for Computational Linguistics (EACL)}, 2023.

\bibitem{lee2022pix2struct}
Kenton Lee, Mandar Joshi, Iulia~Raluca Turc, Hexiang Hu, Fangyu Liu,
  Julian~Martin Eisenschlos, Urvashi Khandelwal, Peter Shaw, Ming-Wei Chang,
  and Kristina Toutanova.
\newblock Pix2struct: Screenshot parsing as pretraining for visual language
  understanding.
\newblock In {\em International Conference on Machine Learning}, pages
  18893--18912. PMLR, 2023.

\bibitem{lewis2006building}
David Lewis, Gady Agam, Shlomo Argamon, Ophir Frieder, David Grossman, and
  Jefferson Heard.
\newblock Building a test collection for complex document information
  processing.
\newblock In {\em Proceedings of the 29th annual international ACM SIGIR
  conference on Research and development in information retrieval}, pages
  665--666, 2006.

\bibitem{li2022dit}
Junlong Li, Yiheng Xu, Tengchao Lv, Lei Cui, Cha Zhang, and Furu Wei.
\newblock Dit: Self-supervised pre-training for document image transformer.
\newblock In {\em Proceedings of the 30th ACM International Conference on
  Multimedia}, pages 3530--3539, 2022.

\bibitem{li2020docbank}
Minghao Li, Yiheng Xu, Lei Cui, Shaohan Huang, Furu Wei, Zhoujun Li, and Ming
  Zhou.
\newblock Docbank: A benchmark dataset for document layout analysis, 2020.

\bibitem{li2021selfdoc}
Peizhao Li, Jiuxiang Gu, Jason Kuen, Vlad~I Morariu, Handong Zhao, Rajiv Jain,
  Varun Manjunatha, and Hongfu Liu.
\newblock Selfdoc: Self-supervised document representation learning.
\newblock In {\em Proceedings of the IEEE/CVF Conference on Computer Vision and
  Pattern Recognition}, pages 5652--5660, 2021.

\bibitem{long2005image}
L~Rodney Long, Sameer~K Antani, and George~R Thoma.
\newblock Image informatics at a national research center.
\newblock {\em Computerized Medical Imaging and Graphics}, 29(2-3):171--193,
  2005.

\bibitem{maini2022augraphy}
Samay Maini, Alexander Groleau, Kok~Wei Chee, Stefan Larson, and Jonathan
  Boarman.
\newblock Augraphy: A data augmentation library for document images.
\newblock {\em arXiv preprint arXiv:2208.14558}, 2022.

\bibitem{mathew2022infographicvqa}
Minesh Mathew, Viraj Bagal, Rub{\`e}n Tito, Dimosthenis Karatzas, Ernest
  Valveny, and CV Jawahar.
\newblock Infographicvqa.
\newblock In {\em Proceedings of the IEEE/CVF Winter Conference on Applications
  of Computer Vision}, pages 1697--1706, 2022.

\bibitem{mathew2020document}
Minesh Mathew, Ruben Tito, Dimosthenis Karatzas, R Manmatha, and CV Jawahar.
\newblock Document visual question answering challenge 2020.
\newblock {\em arXiv preprint arXiv:2008.08899}, 2020.

\bibitem{mungmeeprued2022tab}
Thisanaporn Mungmeeprued, Yuxin Ma, Nisarg Mehta, and Aldo Lipani.
\newblock Tab this folder of documents: page stream segmentation of business
  documents.
\newblock In {\em Proceedings of the 22nd ACM Symposium on Document
  Engineering}, pages 1--10, 2022.

\bibitem{naeini2015obtaining}
Mahdi~Pakdaman Naeini, Gregory Cooper, and Milos Hauskrecht.
\newblock Obtaining well calibrated probabilities using {Bayesian} binning.
\newblock In {\em Proceedings of the AAAI Conference on Artificial
  Intelligence}, volume~29, 2015.

\bibitem{niculescu2005predicting}
Alexandru Niculescu-Mizil and Rich Caruana.
\newblock Predicting good probabilities with supervised learning.
\newblock In {\em Proceedings of the 22nd International Conference on Machine
  learning}, pages 625--632, 2005.

\bibitem{pfitzmann2022doclaynet}
Birgit Pfitzmann, Christoph Auer, Michele Dolfi, Ahmed~S Nassar, and Peter
  Staar.
\newblock Doclaynet: A large human-annotated dataset for document-layout
  segmentation.
\newblock In {\em Proceedings of the 28th ACM SIGKDD Conference on Knowledge
  Discovery and Data Mining}, pages 3743--3751, 2022.

\bibitem{pilan2022text}
Ildik{\'o} Pil{\'a}n, Pierre Lison, Lilja {\O}vrelid, Anthi Papadopoulou, David
  S{\'a}nchez, and Montserrat Batet.
\newblock The text anonymization benchmark (tab): A dedicated corpus and
  evaluation framework for text anonymization.
\newblock {\em Computational Linguistics}, 48(4):1053--1101, 2022.

\bibitem{Powalski2021GoingFB}
Rafal Powalski, Łukasz Borchmann, Dawid Jurkiewicz, Tomasz Dwojak, Michal
  Pietruszka, and Gabriela Pałka.
\newblock Going full-tilt boogie on document understanding with
  text-image-layout transformer.
\newblock In {\em ICDAR}, 2021.

\bibitem{pramanik2020towards}
Subhojeet Pramanik, Shashank Mujumdar, and Hima Patel.
\newblock Towards a multi-modal, multi-task learning based pre-training
  framework for document representation learning.
\newblock {\em arXiv preprint arXiv:2009.14457}, 2020.

\bibitem{russakovsky2015imagenet}
Olga Russakovsky, Jia Deng, Hao Su, Jonathan Krause, Sanjeev Satheesh, Sean Ma,
  Zhiheng Huang, Andrej Karpathy, Aditya Khosla, Michael Bernstein, et~al.
\newblock Imagenet large scale visual recognition challenge.
\newblock {\em International journal of computer vision}, 115:211--252, 2015.

\bibitem{sage2021data}
Cl{\'e}ment Sage, Thibault Douzon, Alex Aussem, V{\'e}ronique Eglin, Haytham
  Elghazel, Stefan Duffner, Christophe Garcia, and J{\'e}r{\'e}my Espinas.
\newblock Data-efficient information extraction from documents with pre-trained
  language models.
\newblock In {\em Document Analysis and Recognition--ICDAR 2021 Workshops:
  Lausanne, Switzerland, September 5--10, 2021, Proceedings, Part II 16}, pages
  455--469. Springer, 2021.

\bibitem{simsa2023docile}
{\v{S}}t{\v{e}}p{\'a}n {\v{S}}imsa, Milan {\v{S}}ulc, Michal
  U{\v{r}}i{\v{c}}{\'a}{\v{r}}, Yash Patel, Ahmed Hamdi, Mat{\v{e}}j
  Koci{\'a}n, Maty{\'a}{\v{s}} Skalick{\`y}, Ji{\v{r}}{\'\i} Matas, Antoine
  Doucet, Micka{\"e}l Coustaty, et~al.
\newblock Docile benchmark for document information localization and
  extraction.
\newblock {\em arXiv preprint arXiv:2302.05658}, 2023.

\bibitem{rossum2022practicalbenchmarks}
{Skalicky, Matyas and Simsa, Stepan and Uricar, Michal and Sulc, Milan}.
\newblock Business document information extraction: Towards practical
  benchmarks, 2022.

\bibitem{kleisterStanislawekGWLK21}
Tomasz Stanislawek, Filip Gralinski, Anna Wr{\'{o}}blewska, Dawid Lipinski,
  Agnieszka Kaliska, Paulina Rosalska, Bartosz Topolski, and Przemyslaw Biecek.
\newblock Kleister: Key information extraction datasets involving long
  documents with complex layouts.
\newblock In {\em ICDAR}, volume 12821 of {\em Lecture Notes in Computer
  Science}, pages 564--579. Springer, 2021.

\bibitem{straydeepform}
J Stray and S Svetlichnaya.
\newblock Deepform: extract information from documents (2020).

\bibitem{tang2023unifying}
Zineng Tang, Ziyi Yang, Guoxin Wang, Yuwei Fang, Yang Liu, Chenguang Zhu,
  Michael Zeng, Cha Zhang, and Mohit Bansal.
\newblock Unifying vision, text, and layout for universal document processing.
\newblock In {\em Proceedings of the IEEE/CVF Conference on Computer Vision and
  Pattern Recognition}, pages 19254--19264, 2023.

\bibitem{tito2022hierarchical}
Rub{\`e}n Tito, Dimosthenis Karatzas, and Ernest Valveny.
\newblock Hierarchical multimodal transformers for multi-page docvqa.
\newblock {\em arXiv preprint arXiv:2212.05935}, 2022.

\bibitem{turski2023ccpdf}
Micha{\l} Turski, Tomasz Stanis{\l}awek, Karol Kaczmarek, Pawe{\l} Dyda, and
  Filip Grali{\'n}ski.
\newblock Ccpdf: Building a high quality corpus for visually rich documents
  from web crawl data.
\newblock {\em arXiv preprint arXiv:2304.14953}, 2023.

\bibitem{dude2023icdar}
Jordy Van~Landeghem, Lukasz Borchmann, Rubèn Tito, Michał Pietruszka, Dawid
  Jurkiewicz, Rafał Powalski, Paweł Józiak, Sanket Biswas, Mickaël
  Coustaty, and Tomasz Stanisławek.
\newblock {ICDAR 2023 Competition on Document UnderstanDing of Everything
  (DUDE)}.
\newblock In {\em Proceedings of ICDAR 2023}, 2023.

\bibitem{vanlandeghem2023document}
Jordy Van~Landeghem, Rub\`{e}n Tito, {\L}ukasz Borchmann, Micha{\l} Pietruszka,
  Pawel Joziak, Rafal Powalski, Dawid Jurkiewicz, Mickael Coustaty, Bertrand
  Ackaert, Ernest Valveny, Matthew~B. Blaschko, Marie-Francine Moens, and
  Tomasz Stanislawek.
\newblock {Document Understanding Dataset and Evaluation (DUDE)}.
\newblock In {\em International Conference on Computer Vision}, 2023.

\bibitem{vapnik1992principles}
Vladimir Vapnik.
\newblock Principles of risk minimization for learning theory.
\newblock In {\em Advances in neural information processing systems}, pages
  831--838, 1992.

\bibitem{vaswani2017attention}
Ashish Vaswani, Noam Shazeer, Niki Parmar, Jakob Uszkoreit, Llion Jones,
  Aidan~N Gomez, {\L}ukasz Kaiser, and Illia Polosukhin.
\newblock Attention is all you need.
\newblock {\em Advances in neural information processing systems}, 30, 2017.

\bibitem{wiedemann2021multi}
Gregor Wiedemann and Gerhard Heyer.
\newblock Multi-modal page stream segmentation with convolutional neural
  networks.
\newblock {\em Language Resources and Evaluation}, 55:127--150, 2021.

\bibitem{yun2021re}
Sangdoo Yun, Seong~Joon Oh, Byeongho Heo, Dongyoon Han, Junsuk Choe, and
  Sanghyuk Chun.
\newblock Re-labeling imagenet: from single to multi-labels, from global to
  localized labels.
\newblock In {\em Proceedings of the IEEE/CVF Conference on Computer Vision and
  Pattern Recognition}, pages 2340--2350, 2021.

\bibitem{zheng2021global}
Xinyi Zheng, Douglas Burdick, Lucian Popa, Xu Zhong, and Nancy Xin~Ru Wang.
\newblock Global table extractor (gte): A framework for joint table
  identification and cell structure recognition using visual context.
\newblock In {\em Proceedings of the IEEE/CVF winter conference on applications
  of computer vision}, pages 697--706, 2021.

\bibitem{zhong2020image}
Xu Zhong, Elaheh ShafieiBavani, and Antonio Jimeno~Yepes.
\newblock Image-based table recognition: data, model, and evaluation.
\newblock In {\em Computer Vision--ECCV 2020: 16th European Conference,
  Glasgow, UK, August 23--28, 2020, Proceedings, Part XXI 16}, pages 564--580.
  Springer, 2020.

\bibitem{zhong2019publaynet}
Xu Zhong, Jianbin Tang, and Antonio~Jimeno Yepes.
\newblock Publaynet: largest dataset ever for document layout analysis.
\newblock In {\em 2019 International Conference on Document Analysis and
  Recognition (ICDAR)}, pages 1015--1022. IEEE, 2019.

\bibitem{Zhu_2022}
Fengbin Zhu, Wenqiang Lei, Fuli Feng, Chao Wang, Haozhou Zhang, and Tat-Seng
  Chua.
\newblock Towards complex document understanding by discrete reasoning.
\newblock In {\em Proceedings of the 30th {ACM} International Conference on
  Multimedia}. {ACM}, oct 2022.

\bibitem{zhu2007automatic}
Guangyu Zhu and David Doermann.
\newblock Automatic document logo detection.
\newblock In {\em Ninth International Conference on Document Analysis and
  Recognition (ICDAR 2007)}, volume~2, pages 864--868. IEEE, 2007.

\end{thebibliography}
}
\clearpage
\newpage
\appendix

\onecolumn

{\noindent \huge Supplementary}

\section{Existing \DC{} datasets}

As the datasets from Table 2 did not satisfy large-scale benchmarking multi-page \DC{} benchmarking requirements, we discuss them in supplementary for interested readers.

\noindent
\textit{Tobacco-3482}  \cite{kumar2013unsupervised} is another subset of IIT-CDIP with fewer samples and a smaller label set than RVL-CDIP.

\jordyskeleton{\noindent
\textit{Tobacco-800} \cite{zhu2007automatic} has been used for page stream segmentation (\cite{wiedemann2021multi}, similarly defined as in \cite{mungmeeprued2022tab}) as it contains consecutively numbered multi-page business documents.}

\noindent
\textit{NIST} The NIST Structured Forms Database~\cite{dimmick1992nist} consists of 5,590 binary synthesized documents from 20 different classes of tax forms. 

\noindent
\textit{MARG} The MARG (Medical Article Records Groundtruth) database~\cite{long2005image} is a layout-based classification benchmark containing 1553 documents which are mainly the first pages of medical journals.

\noindent
\textit{TAB} \cite{mungmeeprued2022tab} is a recently introduced page stream segmentation dataset targeting binary classification to detect document boundaries on multi-page streams. 
It consists of a sample of 44,769 PDF documents from the Truth Tobacco Industry Documents (TTID) archives.

\section{Visualization of proposed \DC{} datasets}\label{app:viz}

As we have contributed two novel datasets consisting of multi-page documents in PDF format, adding visualizations is non-trivial. The datasets are hosted at the HuggingFace Hub (\url{https://huggingface.co/datasets/bdpc}), for which at the time of submission, the dataset viewer does not support PDF data. Rather than adding examples in the manuscript, which is tedious for PDF documents with multiple pages, we have built an interactive app (\url{https://huggingface.co/spaces/jordyvl/viz_bdpc}). This allows for the visualization of samples from the proposed datasets, with an additional filter on the labels, whereas both datasets follow the original RVL-CDIP label taxonomy.

\end{document}